# MeteorologicalSatellite Images Prediction Based on Deep Multi-scales Extrapolation Fusion

Fang Huang[1], Wencong Cheng[1*], PanFeng Wang[1], ZhiGang Wang[1], and HongHong He[1]

*1. Beijing Aviation Meteorological Institute, Beijing, China*

\* Correspondence: emailtocheng@sina.com

**Abstract:** Meteorological satellite imagery is critical for meteorologists. The data have played an important role in monitoring and analyzing weather and climate changes. However, satellite imagery is a kind of observation data and exists a significant time delay when transmitting the data back to Earth. It is important to make accurate predictions for meteorological satellite images, especially the nowcasting prediction up to 2 hours ahead. In recent years, there has been growing interest in the research of nowcasting prediction applications of weather radar images based on deep learning. Compared to the weather radar images prediction problem, the main challenge for meteorological satellite images prediction is the large-scale observation areas and therefore the large sizes of the observation products. Here we present a deep multi-scales extrapolation fusion method, to address the challenge of the meteorological satellite images nowcasting prediction. First, we downsample the original satellite images dataset with large size to several images datasets with smaller resolutions, then we use a deep spatiotemporal sequences prediction method to generate the multi-scales prediction images with different resolutions separately. Second, we fuse the multi-scales prediction results to the targeting prediction images with the original size by a conditional generative adversarial network. The experiments based on the FY-4A meteorological satellite data show that the proposed method can generate realistic prediction images that effectively capture the evolutions of the weather systems in detail. We believe that the general idea of this work can be potentially applied to other spatiotemporal sequence prediction tasks with a large size.

## 1. INTRODUCTION

With the rapid development of satellite remote sensing technology, meteorological satellites play one of the most important roles in atmospheric analysis and forecast. Meteorological satellite data provide an approach to monitor the clouds continuously and track the evolutions of the weather systems. The data are widely used by meteorological researchers and the general public. With the help of meteorological satellite data, researchers can make more accurate weather forecasts. As satellite imagery is a kind of monitoring data, we can only obtain the data in real-time, and due to the limitation of satellite transmission bandwidth and transmission mode, there is a time lag in the collection, sorting, and transmission of the satellite data. In the meanwhile, the weather forecast 0-2 hours ahead in the future, known as nowcasting prediction, is most crucial for meteorologists[1][2]. The operational utility of meteorological satellite data is limited to support the nowcasting needs. Therefore, the accurate prediction for meteorological satellite images, particularly nowcasting prediction up to 2 hours ahead, is crucial to support the real-world socioeconomic needs of many sectors reliant on weather-dependent decision-making.

In this work, we take the meteorological satellite imagery sequences of the past times as the input, trying to generate the satellite images up to 2 hours ahead in the future. This problem is essentially a spatiotemporal sequence forecasting problem. In recent years, Approaches based on deep neural networks have been developed for the problem. The forecast qualities of these methods, as measured by per-grid-cell metrics have greatly improved compared to the traditional prediction methods such as optical flow-based prediction. One recent progress along this path for weather radar images prediction is the convolutional long short term memory(convLSTM) model[3], which extends the fully connected LSTM[4] to set the convolutional structures in both the input-to-state and state-to-state transitions. Wang *et al*.[5][6] extended the convLSTM with zigzag memory flows by Spatiotemporal LSTM(ST-LSTM) model, and then exploit the differential signals to model the non-stationary and approximately stationary properties of the sequence by proposing Memory in Memory (MIM) networks[7]. Generative adversarial networks (GAN) learning [8] has been increasingly used in image generation or prediction [9][10], as it aims to solve the multi-modal training difficulty of the prediction and helps generate less blurry frames. However, the GAN-based methods are weak in time series modeling.

There are two main obstacles when using the existing deep spatiotemporal prediction methods such as ConvLSTM, ST-LSTM, and MIM to predict meteorological satellite images. First, the existing deep spatiotemporal prediction methods require massive computational resources, constraining the input imagery to small sizes. For instance, the size of weather radar imagery was scaled to 100×100 pixels in the work of convLSTM. The sizes of the three datasets (TaxiBJ TrafficFlow, Radar Echo, and Human3.6m) used for the experiments of the MIM model were scaled to 32×32, 64×64, and 128×128 pixels respectively. Under the experimental hardware configuration of this work, we can just process the imagery with the size of 128×128 pixels using the MIM model at the most. On the other hand, the newly launched meteorological satellites generally onboard instruments with high resolutions and large observation regions. For instance, the Advanced Geosynchronous Radiation Imager(AGRI) onboard with Fengyun-4A geostationary meteorological satellite which was launched by China, has a resolution of 500m in the visible light channels, and the resolution of 4km in the infra-red channels. The size of the infra-red channels imagery reaches 1024×1024 pixels even with the 4km resolution when the observation region covers the Northwest Pacific region(10°N-51°N and 90°E-131°E). The sizes will increase when the data cover larger regions and have higher resolutions. So in the normal operational utilities, there are not enough computational resources to predict the meteorological satellite images with large sizes using the existing algorithms directly. According to previous works, we need to split each image with large size into blocks with small size and conduct the prediction on the block sequences separately. Then we combine all the prediction blocks to synthesize the targeting prediction image with the original size. The approach has two disadvantages. First, it tends to miss the evolutions of the weather systems on large scale. To solve images with the size of 1024×1024 pixels, we need to split each imagery sequence into 64(8×8) blocks sequences(the size of each block is 128×128). As shown in Figure 1, the combined prediction images are discontinuous between the adjacent blocks. Second, as some authors have noted[9][10], most deep learning methods can accurately predict low-intensity signals, but their operational utility is limited because their lack of constraints produces blurry images at a longer lead time, yielding poor performance on rarer medium-to-heavy signals. Figure 2 shows the illustrations of the1-hour and 2-hour prediction results of the meteorological satellite images using the MIM[7]. We can find that the prediction images become blurrier with the increased lead time.

Here we present a deep multi-scales extrapolation fusion method for meteorological satellite images nowcasting prediction that mainly addresses the blurry content and the discontinuity between the adjacent blocks of the prediction results. The method can be divided into two phases. In the first phase, to extract the evolutions of the weather systems at different scales, the original satellite images are scaled to several smaller sizes by a down-sampling method. Then the images are composed to imagery sequences with different sizes respectively. Taking the imagery sequences of different sizes as the model input, several deep spatiotemporal prediction models are trained to predict future satellite images on multiple scales. In the second phase, the multi-scales prediction results are fused by a conditional GAN model, to generate the targeting prediction images with the original size. To verify the effectiveness of the proposed method, we conduct experiments on the infra-red 12th channel (bandwidth: 10.3~11.3μm) images of the Feng Yun-4A(FY-4A) geostationary meteorology satellite Advanced Geosynchronous Radiation Imager(AGRI). Experimental results demonstrate that the proposed method can produce realistic prediction satellite images with clear details and continuity, and improve the forecast quality than the corresponding naïve single-scale method.

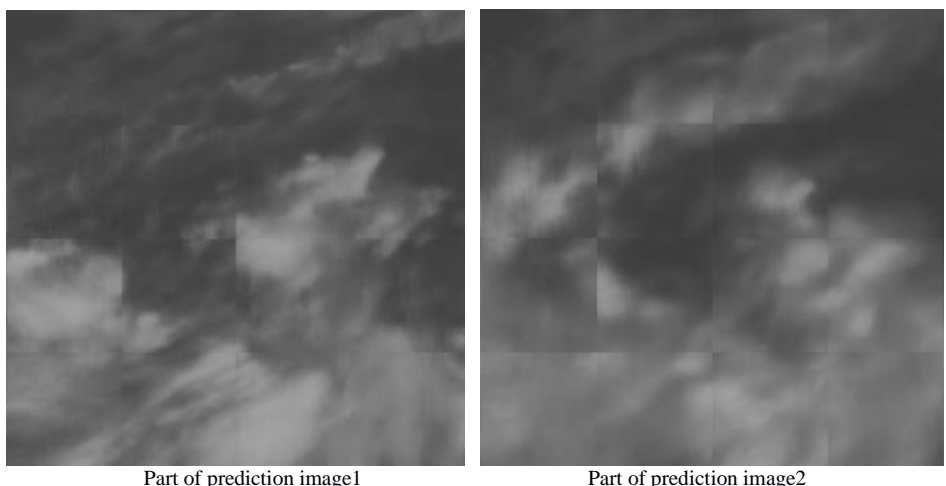

Part of prediction image1　　　　　　　　Part of prediction image2

Figure 1. Discontinuity phenomenon between the adjacent blocks of the prediction images generated by the deep spatiotemporal prediction method (the illustrations are generated by the MIM model)

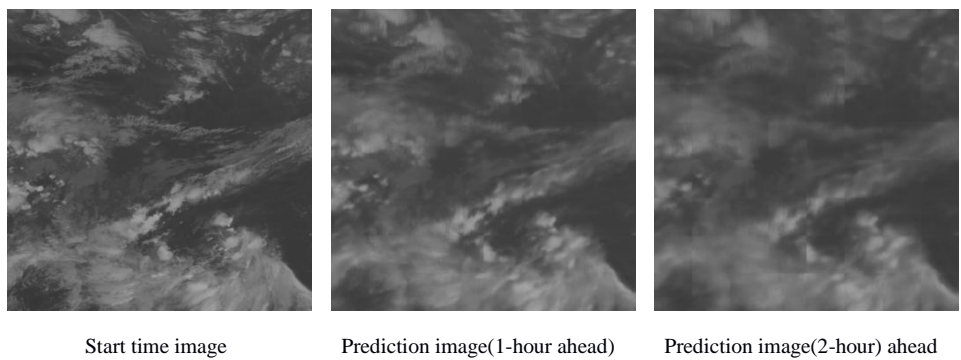

Start time image　　　　Prediction image(1-hour ahead)　　　　Prediction image(2-hour) ahead

Fig.2.Blurry prediction images generated by the deep spatiotemporal prediction method (the illustrations are generated by the MIM model)

**2. Spatiotemporal sequence prediction algorithms based on deep neural network**

Recurrent neural networks(RNNs) are widely used in sequence prediction. In [11], the authors adapted the sequence-to-sequence FC-LSTM framework for multiple frames prediction. Shi, *et al*.[3] [12] extended this model and presented the ConvLSTM by plugging the convolutional operations into recurrent connections. The convolution gates used in the ConvLSTM replace the full connection gate in the FC-LSTM, as shown in Figure 3. A spatiotemporal sequence prediction network using some stacked ConvLSTM units can be presented in Figure 4. The ConvLSTM algorithm is well studied for meteorological radar echoes prediction problem.

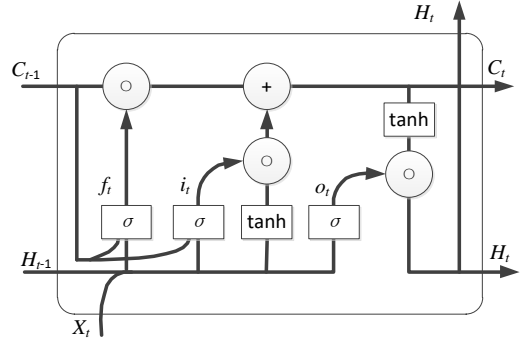

Fig.3. Structure of ConvLSTM unit

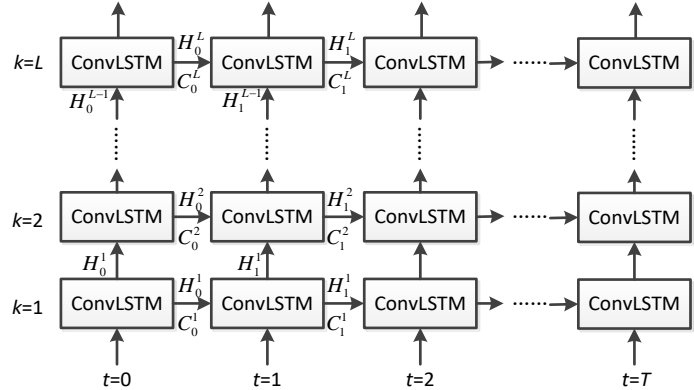

Fig.4 layered ConvLSTM network structure

Let "*" denote convolution operation and "∘" denote Hadamard product of matrix. "$\sigma$" Represents sigmoid function, then convLSTM network unit is defined as follows:

$$i_t = \sigma(W_{xi} * X_t + W_{hi} * H_{t-1} + W_{ci} \circ C_{t-1} + b_i) \quad (1)$$

$$f_t = \sigma(W_{xf} * X_t + W_{hf} * H_{t-1} + W_{cf} \circ C_{t-1} + b_f) \quad (2)$$

$$C_t = f_t \circ C_{t-1} + i_t \circ \tanh(W_{xc} * X_t + W_{hc} * H_{t-1} + b_c) \quad (3)$$

$$o_t = \sigma(W_{xo} * X_t + W_{ho} * H_{t-1} + W_{co} \circ C_t + b_o) \quad (4)$$

$$H_t = o_t \circ \tanh(C_t) \quad (5)$$

There are many improved methods based on the ConvLSTM network. Most of these methods improve the unit structures of the ConvLSTM, enhancing the prediction ability by changing the internal gate connection structure and increasing the time and spatial memory ability, and some of these introduce the short paths to the stacked unit to enhance sequence prediction ability. In this paper, to lay the groundwork for our study, we select the MIM model, one of the latest improved models, as the basic model of sequence prediction. However, it can be replaced by other spatiotemporal sequence prediction models.

## 3. Meteorological satellite images prediction

### 3.1 Data and problem description

This work is conducted on the Fengyun-4A meteorology satellite data. Fengyun-4A meteorological satellite, which is the first satellite of the second generation geostationary meteorological satellites of China, was launched on December 11, 2016[13] [14]. It was fixed at the position of 99.5°E above the equator. The Advanced Geosynchronous Radiation Imager(AGRI) instrument is onboard with the FY-4A satellite as a main observing device. The main task of AGRI is to provide high-frequency, high-precision, and multispectral quantitative remote sensing products of the earth's surface and the cloud systems. Directly serving weather analysis and prediction, short-term climate prediction, and environmental and disaster monitoring. The number of AGRI channels hits 14 (from 0.45μm to 13.8μm) with high spatial and temporal resolutions. Among the 14 channels, there are one 500m resolution channel, two 1km resolution channels, four 2km resolution channels, and seven 4km resolution channels. In this paper, the 12th AGRI channel (infra-red, bandwidth: 10.3~11.3μm) with 4km resolution is selected to conduct the experiments, as it is the most commonly used infra-red channel of FY-4A for meteorologists. We used the hourly sequences as the training and testing dataset.

The goal of this work is to use the previously observed meteorological satellite imagery sequence to forecast the future satellite images up to 2 hours ahead. Suppose we observe an area over a spatial region represented by a $M \times N$ grid which consists of $M$ rows and $N$ columns. A satellite image can be represented by a vector $v \in R^{M \times N}$. The imagery sequence observed continuously can be recorded as $\hat{v}_1, \hat{v}_2, \hat{v}_3, \ldots$. Then the 1-hour ahead prediction problem can be defined to predict the most likely next frame in the future given the previous $j$ images obtained by continuous satellite observation:

$$\tilde{v}_{n+1} = \arg\max p_{(v_{n+1})}(v_{n+1} | \hat{v}_{n-j+1}, \hat{v}_{n-j+2}, \ldots, \hat{v}_n) \quad (6)$$

For the prediction of 2 hours ahead, we use the prediction results of 1-hour ahead to replace the latest real satellite image:

$$\tilde{v}_{n+2} = \arg\max p_{v_{n+2}}(v_{n+2} | \hat{v}_{n-j+2}, \hat{v}_{n-j+3}, \ldots, \tilde{v}_{n+1}) \quad (7)$$

In this work, each input sequence consists of 6 real satellite images at hourly intervals, that is $j = 6$. The models are trained to produce the next 2 images in the sequence, corresponding to the next 2 hours ahead, as shown in Figure 5.

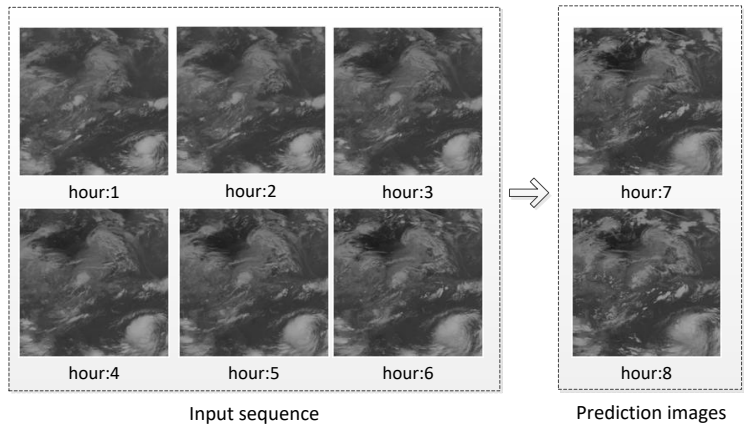

Fig.5 Illustration of meteorological satellite images prediction problem

**3.2 Method**

We now propose a MEF-GAN: a deep Multi-scales Extrapolation Fusion GAN method for meteorological satellite images prediction. MEF-GAN consists of two phases, as shown in Figure 6, a multi-scales extrapolation prediction phase and a results fusion phase. In the first phase, to extract the evolutions of the weather systems on different scales, the satellite images are scaled to several smaller sizes by a down-sampling method. Taking the imagery sequences of different sizes as the input, several deep spatiotemporal prediction models are trained to predict the future satellite images in multiple scales. In the second phase, the multi-scales prediction results gained by the first phase are fused by a conditional GAN to generate the target size prediction images.

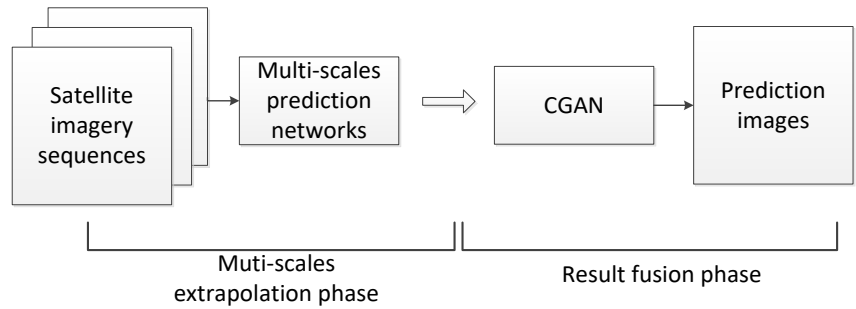

Figure.6 MEF-GAN: a deep Multi-scales Extrapolation Fusion GAN method

In the multi-scales extrapolation prediction phase, we use an average pooling for down-sampling and use 2×2 as the down-sampling coefficient. For instance, if the original size of the images is 1024×1024 pixels, and we set the smallest size as 128×128 pixels, then the sizes of the scaled images after down-sampling are 1024×1024, 512×512, 256×256, and 128×128 pixels respectively. The sequences with the different scales are predicted independently. Due to the constraints of the computational resources, we can only conduct the training for the groundwork spatiotemporal prediction model (*i.e.* MIM model in this paper) with the input size of 128×128 pixels. So the scaled imagery sequences with the size of 128×128 pixels can be used as the model input directly, to extract the whole evolution features of the weather systems. The imagery sequences with the size of 256×256 and above cannot be used as the model input directly, and the images need to be split into blocks with the size of 128×128 pixels. The scaled images with the size of 256×256, 512×512, and 1024×1024 pixels are split into 4 blocks, 16 blocks and 64 blocks respectively. The down-sampling and splitting operations are shown in Figure 7.

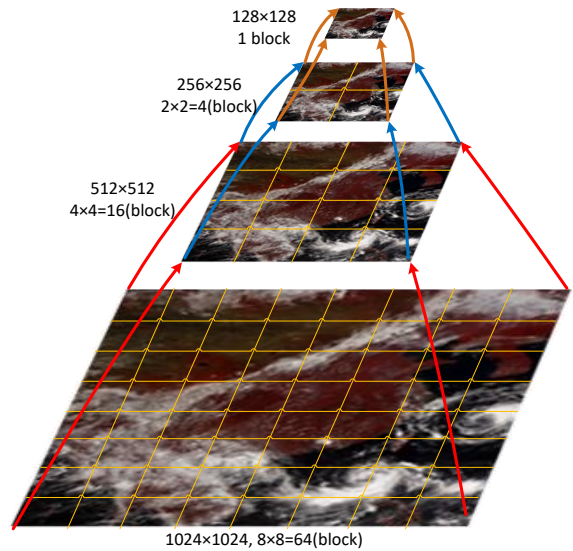

Fig. 7 Down-sampling and splitting operations of a meteorological satellite image

To an imagery sequence with a size larger than 128×128, blocks in a fixed gird position can be formed to a blocks sequence with the size of 128×128 pixels. We can use a trained deep spatiotemporal sequence prediction model to predict the future blocks in that position. Once the blocks in all the grid positions are predicted, we can combine them back to a complete image with the target size. The whole process of the multi-scales extrapolation prediction phase is shown in Figure 8.

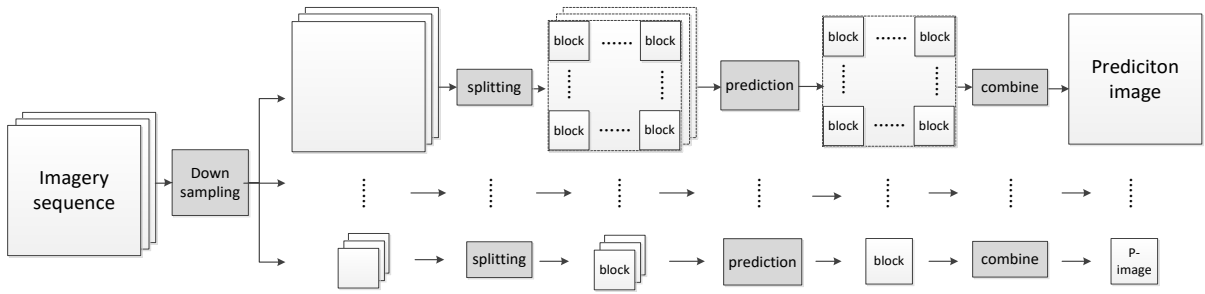
Fig. 8 phase 1:multi-scales extrapolation prediction phase

There are two possible solutions for training the prediction models with the input size larger than 128×128. One solution is to train a single prediction model using all the blocks sequences in every grid-positions to get more training samples. Another solution is to train several independent prediction models for different grid-positions to use the possible existing local information. To compare the two strategies, we conduct additional experiments on the imagery sequences dataset with the size of 256×256 pixels. Each image in the dataset is split into 4 blocks, the experiment results in table 1 show that a single prediction model for all grid-positions is a better choice than training 4 different models for every grid-positions. We think the reason may be the evolutions of the weather systems are similar in different regions, and increasing the train data volume is more important than consideringthe local information. So in the following work, we just train one single prediction model for all the blocks sequences to a certain size without considering the positions of the grids.

In the second phase, *i.e.*, the prediction results fusion phase, we propose to use a CGAN model to create realistic-looking prediction satellite images by fusing the multi-scales prediction results. We constructed the training dataset of phase 2 by reprocessing the training dataset using the multi-scales prediction models of phase 1. The models of phase 1 were run over the training imagery sequences dataset of phase 1, to obtain the multi-scales prediction images of the training dataset. As the real satellite images corresponding to the prediction images are in the training dataset, we can take the multi-scales prediction images and the corresponding real satellite images to compose the training dataset of phase 2.

We train a conditional GAN model to generate the target prediction meteorological satellite images given the corresponding multi-scales prediction images. GANs work by training two different networks: a generator network G and a discriminator network D. G generates the target samples as realistically as possible from the input data. D has trained to estimate the probability of the input drawn from the real data, that is, D tries to classify an input sample as 'real' or 'fake'(synthetic one). Following the GANs principle, both networks are trained simultaneously with D trying to correctly discriminate between real and synthetic samples, while G is trying to produce realistic samples that will confuse D.

For the generator network G, we use a U-Net[15] neural network as the backbone structure. Since the sizes of the most multi-scales prediction images are lower than the original size, an upsampling module is added to adjust the sizes of the prediction images to the original size. D is a standard classification convolution network. D can discern whether the module input data is a real satellite image or a synthetic one. During the GAN training process, D tries to correctly classify the real and the fused prediction satellite images, while G tries to generate the fused prediction satellite images as realistically as possible so that D cannot distinguish between them. To extract the mapping relationship between the input data $x$ and the real satellite data $y$, a CGANmodel is used as the basic structure of D, that is $x$ and $y$ are used together as the input of D (the discriminator of the basic GAN model only uses $y$ as the input).

G and D networks are described in detail in Figure 9. To train the GAN model, we alternatively train the generator network G with one batch of the input data and the discriminator D with two batches, in which one batch contains real samples and the other contains generated samples.

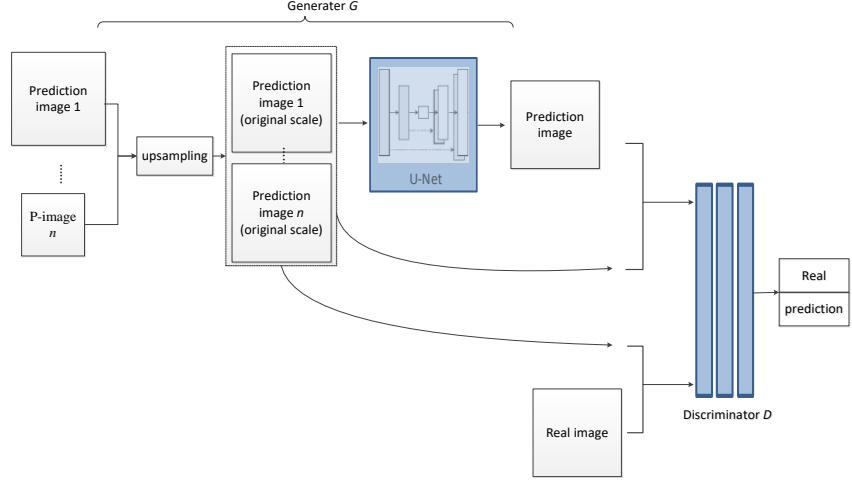

Fig. 9 phase 2: CGAN model in prediction result fusion phase

For D, we would like to find its parameters:

$$argmax_D log(D(x,y)) + log(1 - D(x,G(x,z))) \quad (8)$$

For G, we would like to optimize:

$$argmax_G log(D(x,G(x,z))) \quad (9)$$

The loss function for D is defined as:

$$L_D = L_{bce}(D(x,y),1) + L_{bce}(D(x,G(x,z)),0) \quad (10)$$

where:

$$L_{bce}(\hat{a},a) = -\frac{1}{N}\sum_{i=1}^{N}(a_i log\hat{a}_i + (1-a_i)\log(1-\hat{a}_i)) \quad (11)$$

$N$ is the number of samples in a mini-batch of the model input, $a \in \{0,1\}$ represents the label of the input data (1 for the real satellite data and 0 for the synthetic data), and $\hat{a} \in [0,1]$ is the label estimated by the discriminator D. When the value is close to 0, D assumes the input is the synthetic satellite data, and when the value is close to 1, D assumes the input is the real satellite data.

For G, as mentioned in [16], we use the loss composed of an adversarial term and a reconstruction L1 error. It is defined as:

$$L_G = \lambda_1 L_{bce}(D(x,G(x,z)),1) + \lambda_2 |y - G(x,z)| \quad (12)$$

Where $y$ is the corresponding ground truth satellite data.

**3 Experiment Results and Analysis**

3.1 dataset and experimental environment

To value the effect of the proposed method, we conduct experiments on the No.12 AGRI channels(CH12) of Feng Yun-4A geostationary meteorological satellite, which is an infra-red channel with a 4km resolution. The training and testing data region covers the latitude from 10 °N to 50 °N and longitude from 110 °E to 150 °E, so the size of the original satellite images is 1000×1000 pixels. We scale the images to the size of 1024×1024 for convenience of the following down-sampling operations. The FY-4A satellite data can be downloaded from the website of the National Satellite Meteorological Center (NSMC) of China starting from March 12, 2018. We select the satellite images ranging from March 12, 2018, to June 30, 2019, which are10664 hourly data samples to compose the train dataset. A sliding window with a length of 8 and the step of 1 is used to get imagery sequences from the hourly images. So we can get 10657 training sequences samples with a length of 8. The satellite images ranging from July 1, 2019, to July 31, 2019 are composed of the test dataset, which contains 727 hourly data samples. To ensure the independence of the test samples, every 8 images are composed as a test sequence, forming 88 test sequences without intersection images. We intend to predict the satellite images 2 hours ahead using the previous 6 hours' images. So in each training sequence, the first 6 images are used as the input sequence, and the 7th and 8th images are used as the target labels. Similarly, in each testing sequence, the first 6 images are used as the input sequence, and we try to predict the next 2 images for evaluation.

To value the effect of the proposed method, we use Mean Absolute Error (MAE) and Root Mean Squared Error (RMSE) as the quantitative metrics. In addition, the Peak Signal-to-Noise Ratio (PSNR) and Structural Similarity Index Measure (SSIM), which are commonly used in the domain of image reconstruction and generation, are evaluated. For satellite images with the resolution of $m \times n$, the formulas of the metrics used in this paper are shown as follows:

$$MAE = \frac{1}{mn}\sum_{i=0}^{m-1}\sum_{j=0}^{n-1}|I(i,j) - K(i,j)| \qquad (13)$$

$$RMSE = \sqrt{\frac{1}{mn}\sum_{i=0}^{m-1}\sum_{j=0}^{n-1}[I(i,j) - K(i,j)]^2} \qquad (14)$$

$$PSNR = 10 \times log_{10}\left(\frac{MAX_I^2}{MSE}\right) = 20 \times log_{10}\left(\frac{MAX_I}{\sqrt{MSE}}\right) \qquad (15)$$

$$SSIM(x,y) = \frac{(2\mu_x\mu_y + c_1)(\sigma_{xy} + c_2)}{(\mu_x^2 + \mu_y^2 + c_1)(\sigma_x^2 + \sigma_y^2 + c_2)} \qquad (16)$$

Where $I$ is the prediction result, $K$ is the real satellite image, $MAX_I$ in (15) is the maximum gray scale of the images, which is 255 in this work. In (16), $\mu_x$ and $\mu_y$ are the mean value of $x$ and $y$, $\sigma_x$ and $\sigma_y$ are the standard deviations of $x$ and $y$, $\sigma_{xy}$ is the covariance between $x$ and $y$, and the positive constants $c_1$, $c_2$, and $c_3$ are used to avoid a null denominator.

We implemented the proposed model with TensorFlow 1.6.1 deep learning framework and performed the training using Intel Xeon 4116×2 CPU, 128GB RAM, and NVIDIA TITAN RTX 24GB×2 GPU. The batch size of the prediction models in phase 1 was set to 4 and we trained to models for 150 epochs. The batch size of the fusion GAN model in phase 2 was set to 2 and we trained the model for 300 epochs. The Adam[17] optimizer was used to train the models with a learning rate of 0.002, in which $\beta_1 = 0.5$, $\beta_2 = 0.999$. The training time was approximately 220 hours in phase 1, and 12 hours in phase 2.

As mentioned above, There are two possibilities for training the prediction models with an input size larger than 128×128: we can train a single prediction model using all blocks sequences in every grid-positions and ignore the positions of the blocks to get more training samples (1 model for all grid-positions), or we can train several independent prediction models for different grid-positions (1 model for each grid-position) to extract the possible existed local information of different positions. We conduct experiments on the imagery sequences with the size of 256×256 pixels to compare the two strategies. The images with that size are split into 4 blocks, which the size is 128×128 pixels. We trained the MIM models on the dataset for 150 epochs. The experiment results in table 1 show that the first strategy, *i.e.*, training a single prediction model using all blocks sequences in every grid-positions is a better choice than training several (*i.e.*,4) independent models for different grid-positions.

Table 1 Quantitative comparisons of two different strategies on the dataset with the size of 256×256 pixels

| strategies | MAE | MSE |
|---|---|---|
| 1 model for each grid-position | 5.39 | 61.75 |
| 1 model for all grid-positions | 4.83 | 52.87 |

3.2 Statistical analysis

There are 727 hourly data samples in the test dataset, forming 88 test sequences without intersections. Table 2 and table 3 show the statistical evaluations of 1-hour and 2-hour prediction results for the optical flow method, the MIM (one of the latest deep recurrent neural networks) method, the MIM combined with GAN method(MIM+GAN), and the proposed MEF-GAN method. The MIM method achieves the best accuracies on four metrics in the test, however the prediction results suffer blurry and discontinuous between the adjacent blocks, which are unacceptable in the operational utilities. Both the single-scale method (MIM+GAN) and the multi-scale method (MEF-GAN) achieve better accuracies than the optical flow method, indicating that even though the GAN module introduces additional stochasticity of the prediction results, the deep learning methods have a better cability to extract the correlations of the spatiotemporal sequences compared with the optical flow method. The MEF-GAN method is superior to the single-scale prediction method in the experiments, showing the benefits of extracting features at multiple scales.

Table 2.Quantitative comparisons of different methods for 1-hour prediction

| Methods | MAE | RMSE | PSNR | SSIM |
|---|---|---|---|---|
| MIM | 6.13 | 8.97 | 360.3 | 0.83 |
| Optical flow | 7.30 | 14.24 | 356.3 | 0.78 |
| MIM + GAN | 7.14 | 10.74 | 358.7 | 0.75 |
| MEF-GAN | 6.84 | 10.32 | 359.1 | 0.76 |

Table 3. Quantitative comparisons of different methods for 2-hour prediction

| Methods | MAE | RMSE | PSNR | SSIM |
|---|---|---|---|---|
| MIM | 8.43 | 11.84 | 357.9 | 0.82 |
| Optical flow | 9.20 | 19.34 | 353.7 | 0.75 |
| MIM + GAN | 9.59 | 13.98 | 356.4 | 0.74 |
| MEF-GAN | 9.15 | 13.53 | 356.7 | 0.74 |

3.3 Case study

We select 3 imagery sequences as the study cases. Each input sequence consists of the 6 satellite images with the time interval of 1hour. The sequences are shown in figure 10: Sequence 1. 2019.07.03 00:00~05:00; Sequence 2. 2019.07.03 08:00~13:00; Sequence 3: 2019.07.03 16:00~21:00. We generate 1-hour and 2-hour prediction results by the optical flow method (non-deep learning), the MIM method (naïve deep learning), the MIM with GAN method (single scale) and the MEF-GAN method (multi-scales) respectively. The 1-hour prediction results are shown in Fig.11, which are the prediction results of 2019.07.03 06:00, 14:00 and 22:00. The 2-hour prediction results are shown in Fig.12, which are the prediction results of 2019.07.03 07:00, 15:00 and 23:00.

The prediction images generated by the proposed method are clear in detail and continuous between the adjacent blocks, and are realistic-looking to the real satellite images. The prediction images generated by the optical flow method have hollow spaces near the borders and have several distorted parts. The prediction images generated by the MIM model are discontinuous between the adjacent blocks due to the results being stitched by several small prediction blocks, and the prediction results are increasingly blurry with the longer leading time.

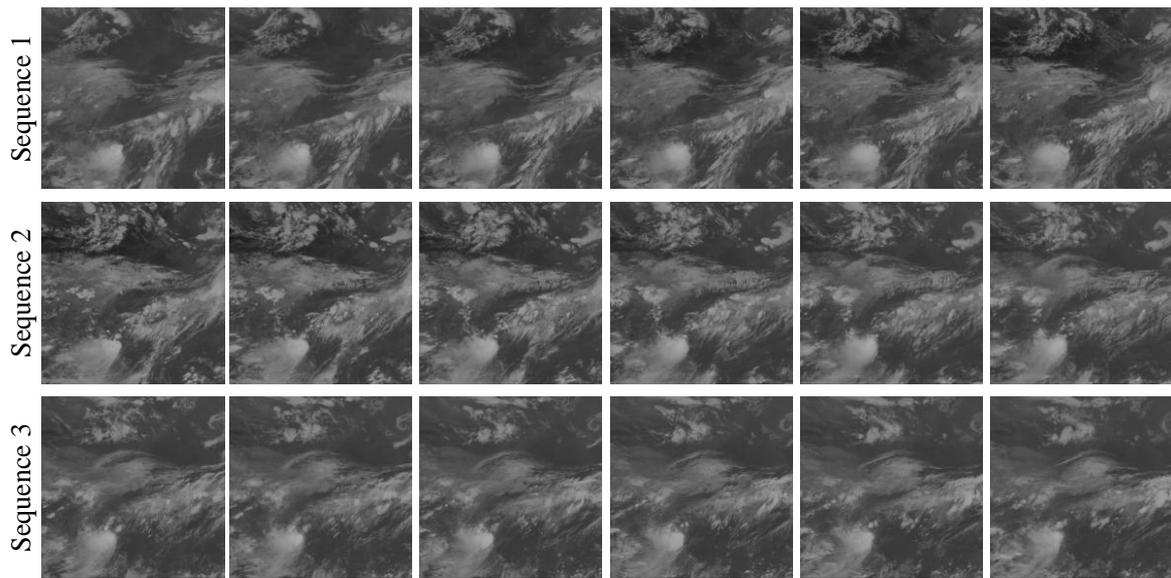

Fig.10 Study cases of the input sequences(Sequence 1: 2019.07.03 00:00~05:00; Sequence 2: 2019.07.03 08:00~13:00; Sequence 3: 2019.07.03 16:00~21:00)

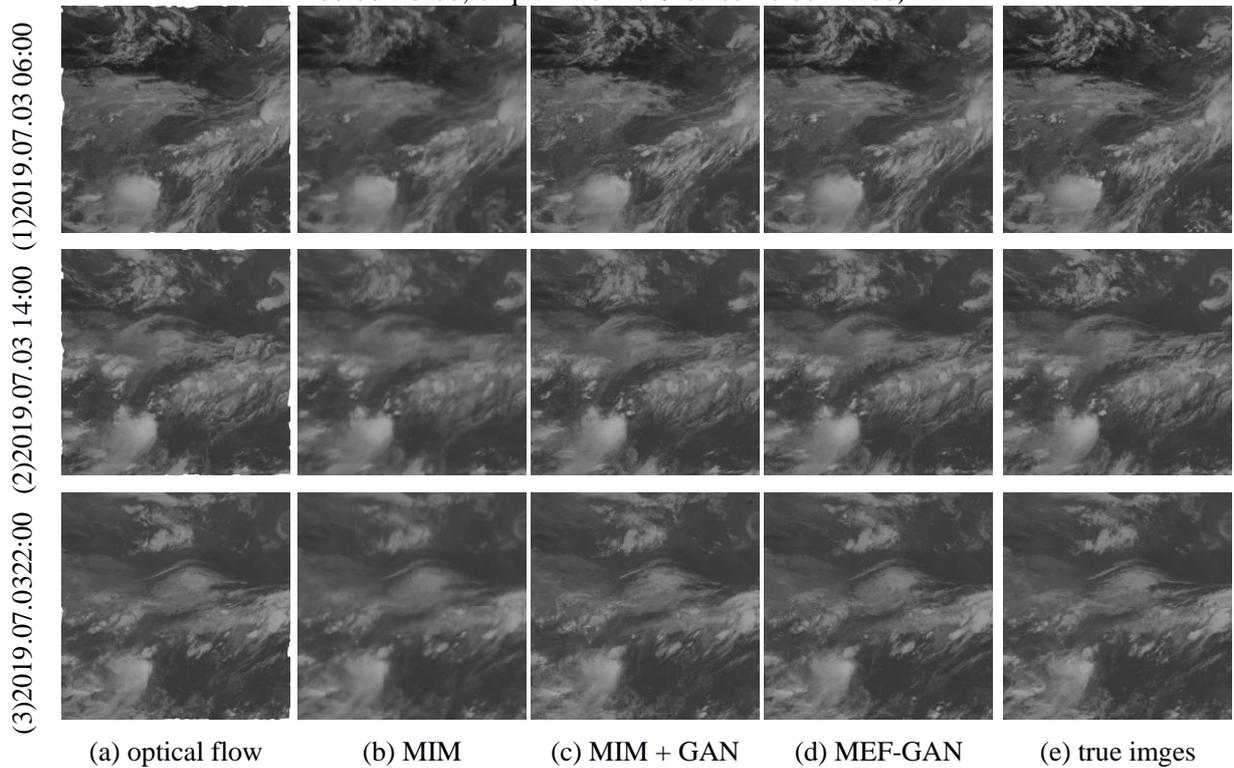

(a) optical flow     (b) MIM     (c) MIM + GAN     (d) MEF-GAN     (e) true imges

Fig.11 Results of the study cases on the challenging methods (1-hour prediction results)

The quantitative evaluations of the 3 study cases are shown in Table 4 to Table 9. The conclusions are mainly as the same as the statistical analysis. The MIM method achieves the best accuracies on the four metrics, however the prediction images suffer the blurry content and discontinuity between the adjacent blocks. The performances of the MIM+GAN (single scale) method are superior in RMSE and PSNR metrics and slightly inferior in MAE and SSIM metrics to the optical flow method, due to the GAN module introducing additional stochasticity to gain the realistic-looking results. The MEF-GAN (multi-scales) method achieves better performance than the optical flow method and the MIM+GAN method. Although the quantitative performances of the MEF-GAN are inferior to the naïve MIM method, the perceptual feelings of the results are much better than the optical flow method and the MIM method. The quantitative evaluations of the prediction results also verify the effectiveness of the proposed method.

Table 4 Quantitative comparisons of different methods for 1-hour prediction at 06:00

| Methods | MAE | RMSE | PSNR | SSIM |
|---|---|---|---|---|
| MIM | 7.25 | 10.83 | 358.6 | 0.82 |
| Optical flow | 8.85 | 16.24 | 355.1 | 0.68 |
| MIM + GAN | 8.99 | 13.29 | 356.8 | 0.64 |
| MEF-GAN | 8.03 | 12.05 | 357.7 | 0.67 |

Table 5 Quantitative comparisons of different methods for 1-hour prediction at 14:00

| Methods | MAE | RMSE | PSNR | SSIM |
|---|---|---|---|---|
| MIM | 5.99 | 8.91 | 360.3 | 0.87 |
| Optical flow | 7.84 | 14.63 | 355.9 | 0.73 |
| MIM + GAN | 7.09 | 10.70 | 358.7 | 0.72 |
| MEF-GAN | 6.91 | 10.29 | 359.0 | 0.72 |

Table 6 Quantitative comparisons of different methods for 1-hour prediction at 22:00

| Methods | MAE | RMSE | PSNR | SSIM |
|---|---|---|---|---|
| MIM | 5.02 | 7.60 | 361.7 | 0.90 |
| Optical flow | 6.14 | 11.90 | 357.8 | 0.78 |
| Deep+ GAN | 6.16 | 9.43 | 359.8 | 0.76 |
| MEF-GAN | 5.86 | 8.90 | 360.3 | 0.76 |

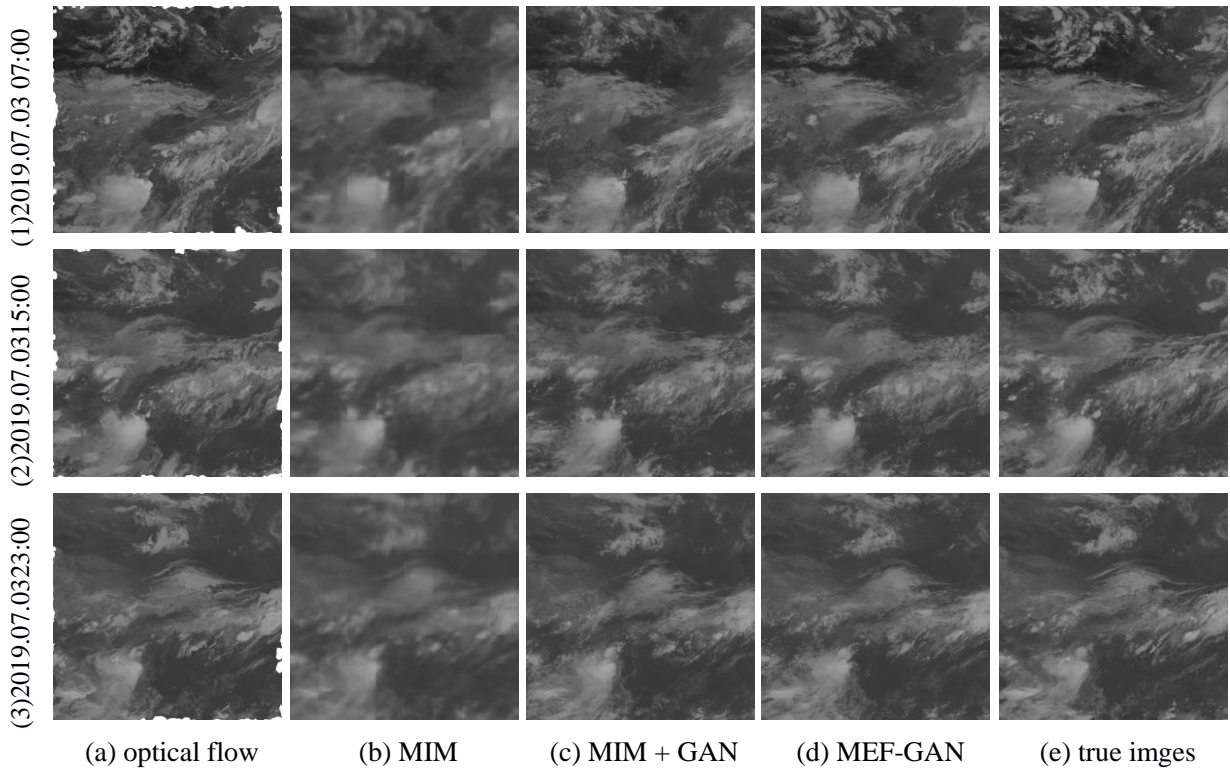

(a) optical flow   (b) MIM   (c) MIM + GAN   (d) MEF-GAN   (e) true imges

Fig.12 Results of the study cases on the challenging methods (2-hour prediction results)

Table 7 Quantitative comparisons of different methods for 2-hour prediction at 07:00

| Methods | MAE | RMSE | PSNR | SSIM |
|---|---|---|---|---|
| MIM | 11.11 | 15.42 | 355.5 | 0.80 |
| Optical flow | 11.91 | 23.26 | 352.0 | 0.72 |
| Deep+ GAN | 12.08 | 17.31 | 354.5 | 0.72 |
| MEF-GAN | 11.31 | 16.57 | 354.9 | 0.72 |

Table 8 Quantitative comparisons of different methods for 2-hour prediction at 15:00

| Methods | MAE | RMSE | PSNR | SSIM |
|---|---|---|---|---|
| MIM | 8.32 | 11.62 | 358.0 | 0.86 |
| Optical flow | 9.35 | 19.79 | 353.4 | 0.80 |
| Deep+ GAN | 9.56 | 13.95 | 356.4 | 0.79 |
| MEF-GAN | 8.91 | 13.17 | 356.9 | 0.79 |

Table 9 Quantitative comparisons of different methods for 2-hour prediction at 23:00

| Methods | MAE | RMSE | PSNR | SSIM |
|---|---|---|---|---|
| MIM | 7.60 | 10.49 | 358.9 | 0.88 |
| Optical flow | 7.80 | 17.93 | 354.2 | 0.83 |
| Deep+ GAN | 8.38 | 12.43 | 357.4 | 0.82 |
| MEF-GAN | 7.74 | 11.56 | 358.0 | 0.82 |

**4 conclusions**

In this work, we study the meteorological satellite images prediction problem by introducing a deep multi-scale extrapolation fusion method: MEF-GAN. The proposed method first predict the evolutions of the weather systems at different scales by training several deep spatiotemporal sequence prediction models, and then we fuse the multi-scales prediction results to the targeting prediction images with the original size by a conditional generative adversarial network. The method can effectively predict meteorological satellite images with large size and can generate realistic-looking images. The experiments based on the FY-4A meteorological satellite data show that the proposed method can generate realistic prediction images that are effective to capture the evolutions of the weather systems in detail. We believe that the general idea of this work can be potentially applied to other spatiotemporal sequence prediction tasks with a large size.

**Reference**

[1] J. W. Wilson, Y. Feng, M. Chen, and R. D. Roberts, "Nowcasting Challenges during the Beijing Olympics: Successes, Failures, and Implications for Future Nowcasting Systems," *Weather Forecast.*, vol. 25, no. 6, pp. 1691–1714, 2010.

[2] W. Yong, E. D. Coning, A. Harou, W. Jacobs, and J. Sun, *Guidelines for Nowcasting Techniques*, vol. 1198. World Meteorological Organization, 2017.

[3] X. Shi, Z. Chen, H. Wang, D.-Y. Yeung, W. Wong, and W. Woo, "Convolutional LSTM Network: A Machine Learning Approach for Precipitation Nowcasting," in *Proceedings of the 28th International Conference on Neural Information Processing Systems - Volume 1*, Cambridge, MA, USA, 2015, pp. 802–810.

[4] S. Hochreiter and J. Schmidhuber, "Long Short-term Memory," *Neural Comput.*, vol. 9, pp. 1735–80, Dec. 1997, doi: 10.1162/neco.1997.9.8.1735.

[5] Y. Wang, M. Long, J. Wang, Z. Gao, and P. S. Yu, "PredRNN: Recurrent Neural Networks for Predictive Learning Using Spatiotemporal LSTMs," in *Proceedings of the 31st International Conference on Neural Information Processing Systems*, Red Hook, NY, USA, 2017, pp. 879–888.

[6] Y. Wang, Z. Gao, M. Long, J. Wang, and P. S. Yu, "PredRNN++: Towards A Resolution of the Deep-in-Time Dilemma in Spatiotemporal Predictive Learning," in *Proceedings of the 35th International Conference on Machine Learning*, Jul. 2018, vol. 80, pp. 5123–5132. [Online]. Available: https://proceedings.mlr.press/v80/wang18b.html

[7] Y. Wang, J. Zhang, H. Zhu, M. Long, and P. S. Yu, "Memory in Memory: A Predictive Neural Network for Learning Higher-Order Non-Stationarity From Spatiotemporal Dynamics," 2019.

[8] I. J. Goodfellow *et al.*, "Generative Adversarial Nets," in *Proceedings of the 27th International Conference on Neural Information Processing Systems - Volume 2*, Cambridge, MA, USA, 2014, pp. 2672–2680.

[9] Z. Xu, J. Du, J. Wang, C. Jiang, and Y. Ren, "Satellite Image Prediction Relying on GAN and LSTM Neural Networks," in *ICC 2019 - 2019 IEEE International Conference on Communications (ICC)*, 2019, pp. 1–6. doi: 10.1109/ICC.2019.8761462.

[10] S. Ravuri *et al.*, "Skillful Precipitation Nowcasting using Deep Generative Models of Radar," *Nature*, vol. 597, no. 12, pp. 672–677, 2021.

[11] N. Srivastava, E. Mansimov, and R. Salakhutdinov, "Unsupervised Learning of Video Representations Using LSTMs," in *Proceedings of the 32nd International Conference on International Conference on Machine Learning - Volume 37*, Lille, France, 2015, pp. 843–852.

[12] X. Shi *et al.*, "Deep Learning for Precipitation Nowcasting: A Benchmark and a New Model," in *Proceedings of the 31st International Conference on Neural Information Processing Systems*, Red Hook, NY, USA, 2017, pp. 5622–5632.

[13] P. Zhang, L. Chen, D. Xian, and Z. Xu, "Recent progress of Fengyun meteorology satellites," *Chin. J. Space Sci.*, vol. 38, no. 5, pp. 788–796, Sep. 2018, doi: 10.11728/cjss2018.05.788.

[14] P. Zhang *et al.*, "Latest Progress of the Chinese Meteorological Satellite Program and Core Data Processing Technologies," *Adv. Atmospheric Sci.*, vol. 36, no. 9, pp. 1027–1045, Sep. 2019, doi: 10.1007/s00376-019-8215-x.

[15] O. Ronneberger, P. Fischer, and T. Brox, "U-Net: Convolutional Networks for Biomedical Image Segmentation," in *International Conference on Medical Image Computing and Computer-Assisted Intervention*, Oct. 2015, vol. 9351, pp. 234–241. doi: 10.1007/978-3-319-24574-4_28.

[16] D. Pathak, P. Krahenbuhl, J. Donahue, T. Darrell, and A. A. Efros, "Context Encoders: Feature Learning by Inpainting," in *2016 IEEE Conference on Computer Vision and Pattern Recognition (CVPR)*, Los Alamitos, CA, USA, 2016, pp. 2536–2544. doi: 10.1109/CVPR.2016.278.

[17] D. Kingma and J. Ba, "Adam: A Method for Stochastic Optimization," *Int. Conf. Learn. Represent.*, Dec. 2014.